\documentclass[letterpaper, 10 pt, conference]{ieeeconf}  % Comment this line out if you need a4paper

\IEEEoverridecommandlockouts                              % This command is only needed if 
\usepackage{graphics} % for pdf, bitmapped graphics files
\usepackage{graphicx}
\usepackage{hyperref}
\usepackage{algorithm}
\usepackage{amsmath}
\usepackage[noend]{algpseudocode}

\title{\LARGE \bf
Autonomous navigation for low-altitude UAVs in urban areas
}

\author
{
	Thomas Castelli$^{1,2}$, Aidean Sharghi$^{3}$, Don Harper$^{3}$, Alain Tremeau$^{2}$ and Mubarak Shah$^{3}$% <-this % stops a space
\thanks{$^{1}$ Survey Copter / Airbus Defense and Space, Pierrelatte, FRANCE}
\thanks{$^{2}$ Hubert Curien Laboratory, Saint-Etienne, FRANCE%
	{\tt\small \newline alain.tremeau@univ-st-etienne.fr,\newline thomas.castelli@univ-st-etienne.fr}}%
\thanks{$^{3}$ Center for Research in Computer Vision, UCF, Orlando, USA
	    {\tt\small \newline aidean.sharghi@knights.ucf.edu, harper@ucf.edu, shah@crcv.ucf.edu}}%
}

\begin{document}

\maketitle
\thispagestyle{empty}
\pagestyle{empty}

%%%%%%%%%%%%%%%%%%%%%%%%%%%%%%%%%%%%%%%%%%%%%%%%%%%%%%%%%%%%%%%%%%%%%%%%%%%%%%%%
\begin{abstract}

In recent years, consumer Unmanned Aerial Vehicles have become very popular, everyone can buy and fly a drone without previous experience, which raises concern in regards to regulations and public safety. In this paper, we present a novel approach towards enabling safe operation of such vehicles in urban areas. \linebreak
Our method uses geodetically accurate dataset images with Geographical Information System (GIS) data of road networks and buildings provided by Google Maps, to compute a weighted A* shortest path from start to end locations of a mission. Weights represent the potential risk of injuries for individuals in all categories of land-use, i.e. flying over buildings is considered safer than above roads. We enable safe UAV operation in regards to 1- land-use by computing a static global path dependent on environmental structures, and 2- avoiding flying over moving objects such as cars and pedestrians by dynamically optimizing the path locally during the flight. As all input sources are first geo-registered, pixels and GPS coordinates are equivalent, it therefore allows us to generate an automated and user-friendly mission with GPS waypoints readable by consumer drones' autopilots.
We simulated 54 missions and show significant improvement in maximizing UAV's standoff distance to moving objects with a quantified safety parameter over 40 times better than the naive straight line navigation.
\end{abstract}

%%%%%%%%%%%%%%%%%%%%%%%%%%%%%%%%%%%%%%%%%%%%%%%%%%%%%%%%%%%%%%%%%%%%%%%%%%%%%%%%
\section{INTRODUCTION}
UAVs are becoming increasingly present in our everyday lives, their extensive use recently jumped from military to hobby and professional applications. The consumer market is growing and now it offers a wide range of micro and mini UAVs at affordable costs. But this popularity induces some dangerous behavior, most people do not realize that a simple mistake can cause severe injuries to themselves or others.\linebreak
In the United-States, the FAA has taken measures to inform hobbyists and encourage them to follow a code of conduct to prevent accidents. The only form available is the advisory circular \textquoteleft AC 91-57 \textquoteright from June 9th 1981, it advises pilots to keep their UAVs within their line of sight, below 400 feet above ground level, further than 5 miles from an airport (or warn them), and to avoid flying above people.\linebreak
Even for the vast majority of UAV users that are responsible and careful in their use, there is no automated means to fly safely in regards to the UAV's environment. Our work aims to provide such functionality to micro and mini UAVs that are operated in urban areas.

\begin{figure}[thpb]
	\centering
	\includegraphics[width=1\linewidth,height=\textheight,keepaspectratio]{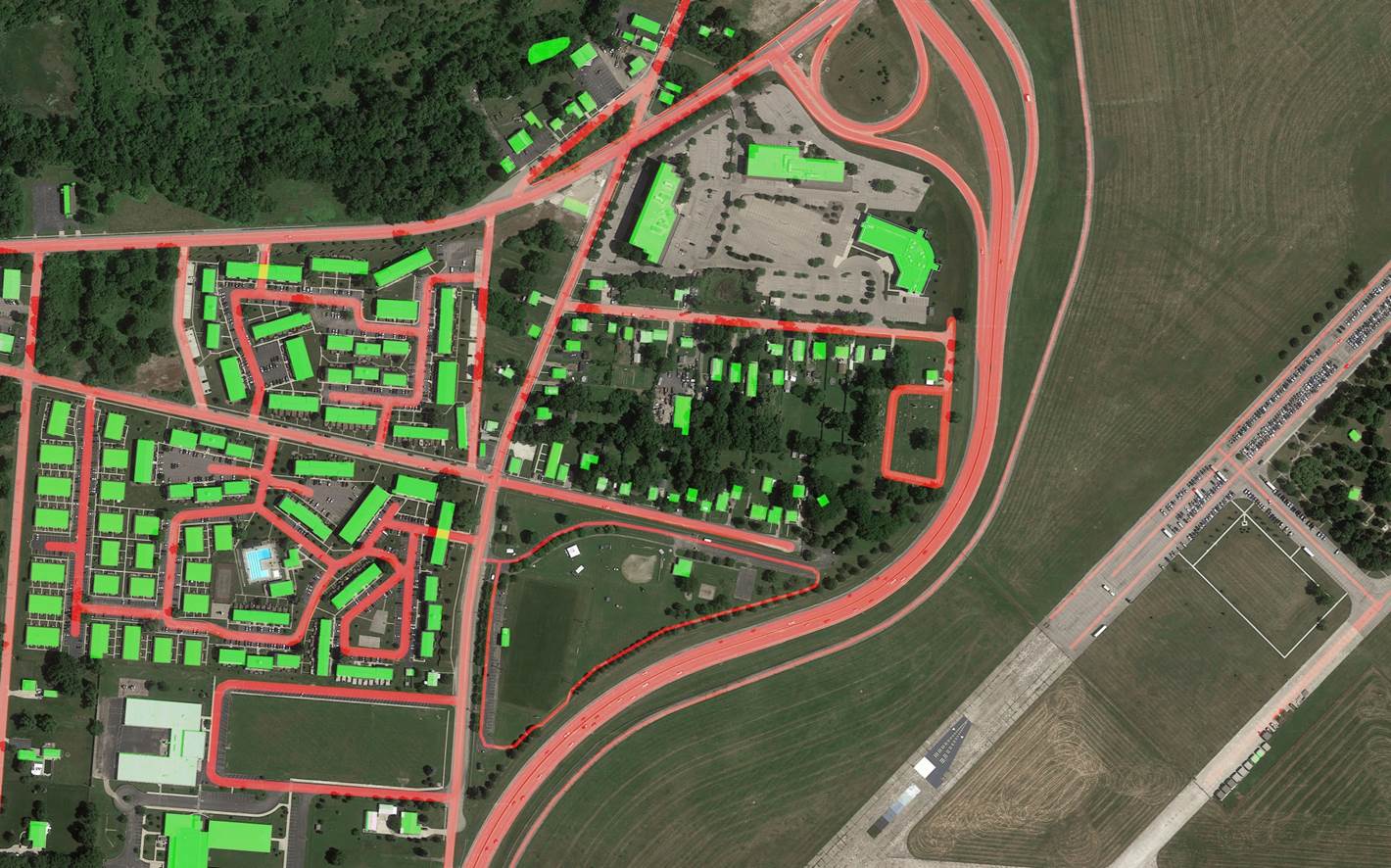}
	\caption{ Visualization of the weight map overlaid on the corresponding satellite image, for WPAFB dataset. Colors
		represent costs in the weight map, red, transparent and green respectively represent dangerous, neutral, and safer areas. 
		%Visualization of the weight map on top of the corresponding satellite image, for WPAFB dataset, using GIS for global path planning. Colors represent costs in the weight map, red, transparent and green represent respectively dangerous, neutral, and safer areas.
		}
	\label{map}
\end{figure}

In this paper, we propose a novel method for autonomous navigation for low-altitude UAVs in urban areas. For a given mission our method computes safe waypoints, which dynamically adapt the flight plan to the UAV’s surroundings by avoiding objects such as cars and pedestrians. We take advantage of satellite and georegistered data to adapt the UAV’s mission layout by computing a weighted shortest path instead of flying in a straight line. Weights in our cost function for computing the flight path are defined using land-use summarized in three classes: most dangerous areas are roads and paths where people are prone to the danger the UAV represents, safest are buildings and water, and the rest is in between (Fig. \ref{map}). For increased safety, our method also adapts dynamically to moving objects while in flight by adding new local weight to the global weight map.

In our general scenario, we assume a UAV with video camera flying over a given geographical region, for which geodetically accurate reference image, GIS data of buildings and road networks are available. Captured videos are  georegistered with the reference image in order to transfer pixel coordinates to GPS coordinates.  Moving objects, e.g. vehicles and pedestrians, are detected and tracked from frame to frame. Given the tracks, GIS and reference image, the optimal UAV path is dynamically computed. For simplicity in this paper, we employ ground truth tracks available from WPAFB and PVLabs datasets providing geo-registered images and
ground truth for moving objects. Finally, we simulate a real flight by complying with the \textquoteleft AC 91-57\textquoteright form and using parameters of compatible hardware.

%Our proposed method enables the drone operator to include a safety step in the waypoints definition for his mission by adapting the flight plan to the UAV's surroundings, and also during the flight by avoiding objects such as cars and pedestrians. We are taking advantage of satellite and geo-registered data to adapt the UAV's mission layout by computing a weighted shortest path instead of flying a straight line. Weights are defined using land-use summarized in three classes: most dangerous areas are roads and paths where people are prone to the danger the UAV represents, safest are buildings and water, and the rest is in between (Fig. \ref{map}). For increased safety, our method also adapts dynamically to moving objects while in flight by adding new local weights to the global weight map. Experiments are conducted in WPAFB and PVLabs dataset providing geo-registered images and ground truth for moving objects, we simulate a real flight by complying with the "AC 91-57" form and using parameters of compatible hardware.

%%%%%%%%%%%%%%%%%%%%%%%%%%%%%%%%%%%%%%%%%%%%%%%%%%%%%%%%%%%%%%%%%%%%%%%%%%%%%%%%
\section{RELATED WORK}
Many different topics are studied to enhance the usability and to develop new functions to make drones more capable and autonomous. There are several subfields which are related to this work including video geo-registration, detection and tracking of moving objects in videos, detection of roads, buildings, water bodies from satellite imagery and flight path planning. 
%Because of their growing popularity in regards to new applications and fields of research, recent work are very diverse. 

The most popular trend in UAV video analysis has been moving object detection and tracking from aerial images, many approaches have been proposed with or without using GIS data and geo-registration steps. Kimura \textit{et al.} \cite{MODAT1} use epipolar constraint and flow vector bound to detect moving objects, Teutsch \textit{et al.} \cite{MODAT2} employ explicit segmentation of images, Xiao \textit{et al.} \cite{MODAT3} restrain the search on the road network, and Lin \textit{et al.} \cite{MODATGEO1} use a motion model in geo-coordinates. Moving object detection and tracking are mainly used to follow targets, for surveillance as Quigley \textit{et al.} \cite{target} and Rafi \textit{et al.} \cite{auto} describe with their flight path adaptation solutions, or for consumer applications at very low-altitude as in \cite{FOLLOW1} and \cite{FOLLOW2}.

%On a broader domain, DPA* \cite{short1}, presents an approach to prune the search space based on previous path and its relation to the dynamic event. Xu et. al \cite{short4} aims to find the optimal path efficiently in the octree-partitioned search space, designing a planner for MAVs. 

Another area that has been getting a lot of attention is autonomous navigation. Different subproblems have been studied, path planning in dynamic environment \cite{short1, short4}, GIS-assisted and vision-based localization using either road detection \cite{LOC1}, buildings layout \cite{GEOREG2} or DEM (Digital Elevation Map) \cite{GEOREG1}. Various methods have been proposed for UAV navigation, using optical flow with \cite{NAV1} or without DEM \cite{NAV2}, or using inertial sensors \cite{IMU1}.\\ 
Obstacle avoidance is also a big concern for automating UAV operation, but research has mostly been focused on ground robots \cite{c3, obstacle2}, even if there has been adaptations for UAVs as Israelsen \textit{et al.}'s intuitive solution for operators \cite{obstacle1}.

The approaches for autonomously navigating UAVs have been studied, but previous work focus on target following or keeping the UAV's integrity. However, in this paper we propose an autonomous UAV navigation method in order to increase public safety in regards to drones operation, and also to prevent UAVs finding themselves in difficult situations.
%our approach aims to provide a more global step to first increase public safety in regards to drones operation, and second to prevent UAVs finding themselves in difficult situations.

%%%%%%%%%%%%%%%%%%%%%%%%%%%%%%%%%%%%%%%%%%%%%%%%%%%%%%%%%%%%%%%%%%%%%%%%%%%%%%%%
\section{OUR METHOD}
Our contribution towards safe integration of small UAVs into the airspace has two main steps. The first step, described in section A and B, takes into account the physical surroundings of the UAV by computing, as part of the mission preparation, a global path between the user-given start and end locations. This path is represented as a succession of waypoints, exactly as users are accustomed to in mission planner softwares. Before takeoff the user is able to validate the automated path, and he can modify the waypoints if needed.
The second step, described in section C, runs in online fashion during the flight and takes into account the environment of the UAV by dynamically adapting its behavior in regards to moving objects that need to be avoided.

\subsection{Extracting the geo-referenced weight map} \label{sec3A}
A convenient approach to gain awareness of the UAV's surroundings is to use satellite imagery and the meta-data provided by \href{https://www.google.com/maps}{Google Maps}, \href{https://www.digitalglobe.com/}{DigitalGlobe}, \href{https://www.planet.com/}{Planet Labs} or others.
To jointly use geo-registered data and aerial imagery obtained from UAV, there has to be a common representation and space. Public tools providing satellite images are very popular and well integrated in third party UAV software such as \href{http://planner.ardupilot.com/}{Mission Planner}. For simplicity and compliance, the solution is then to register video images onto a geo-registered satellite image of the area of interest. UAV and world coordinates systems are related with \eqref{sensor}, as described in \cite{c7}. %using a sensor model $\vec X_{camera}$ depending on camera pose relative to the UAV and world coordinates , .

\vspace{-0.02in}
\begin{equation}
\vec X_{camera} = G_{y}G_{z}R_{y}R_{x}R_{z}T\vec X_{world}
\label{sensor}
\end{equation} 
with $\vec X_{world}$ the world coordinates system, $T$ is the translation matrix derived from the vehicle's latitude, longitude and altitude, and $G_{y}$, $G_{z}$, $R_{y}$, $R_{x}$ and $R_{z}$ are rotation matrices regarding respectively camera elevation angle, camera scan angle, vehicle pitch angle, vehicle roll angle and vehicle heading angle.\\

We chose to use Google Maps API for it's convenience and for the quality of the data provided\footnote{The proposed method is not dependent on the source, any satellite image and data provider can be used.}. This free API allows anyone to request satellite images and roadmaps displaying buildings and roads. These three links,  \href{http://maps.googleapis.com/maps/api/staticmap?center=28.600558,-81.197722&zoom=18&format=png32&sensor=false&size=640x640&scale=2&maptype=satellite&style=feature:poi|visibility:off&style=feature:road|element:labels|visibility:off&style=feature:transit|visibility:off&style=feature:landscape|element:labels|visibility:off&style=element:labels|visibility:off&style=feature:landscape|element:geometry.stroke|visibility:off&style=feature:poi|element:geometry.stroke|visibility:off}{1},  \href{http://maps.googleapis.com/maps/api/staticmap?center=28.600558,-81.197722&zoom=18&format=png32&sensor=false&size=640x640&scale=2&maptype=roadmap&style=visibility:off&style=feature:road|visibility:on&style=element:labels|visibility:off&style=element:geometry|color:0x000000}{2} and \href{http://maps.googleapis.com/maps/api/staticmap?center=28.600558,-81.197722&zoom=18&format=png32&sensor=false&size=640x640&scale=2&maptype=roadmap&style=visibility:off&style=feature:landscape|visibility:on&style=feature:landscape|element:labels|visibility:off&style=feature:water|visibility:on|color:0x000000&style=feature:landscape.natural|visibility:off}{3}, give example commands to request satellite, road map and building images.

We assume that flying above buildings represents less risk than doing so above other environmental elements such as roads or crowded streets. The resulting weight map for the Wright-Patterson Air Force Base (WPAFB) area, shown in Fig. \ref{map}, displays three categories: 
\begin{itemize}
\item Red is to be avoided, for roads and paths. 
\item Green is to be preferred, for buildings and water.
\item Transparent is in between, for other land-use.\newline
\end{itemize}

\begin{figure}[thpb]
	\centering
	\includegraphics[width=1.15\linewidth,height=\textheight,keepaspectratio]{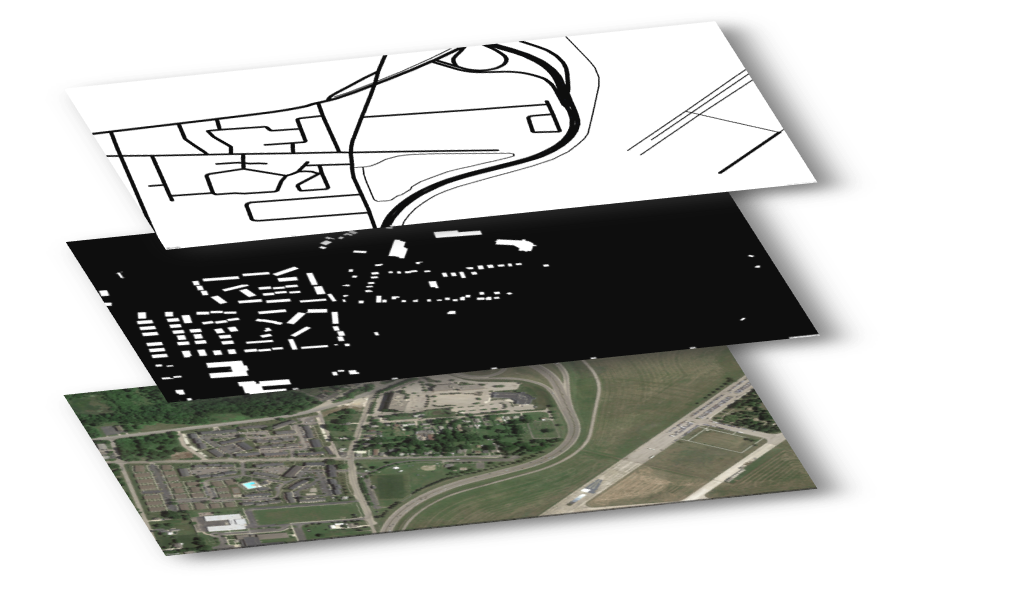}
	\caption{This figure shows from bottom to top three types of data used in this work: satellite image, buildings and water map, and roads map.}
	\label{layers}
\end{figure}

To extract the map, only two GPS locations are needed as input from the operator: top left and bottom right GPS coordinates. A grid of image GPS locations is then computed based on Google Maps' camera parameters and resolution level, in other words the ground sampling distance (GSD), to ensure sufficient overlap between images for stitching. We have defined the GPS grid in a way that successive images have pure translations between them. We thus can stitch them together using straight forward normalized cross correlation, which is a robust and fast method given that we manipulate large images and avoid scale change and rotation. This process allows us to minimize the error while creating the geo-registered map. As a result we obtain 3 images for any given area (Fig. \ref{layers}). Given the image center GPS location, $Lat_{1}$ and $Lon_{1}$, the corresponding GPS location ($Lat_{2}$ and $Lon_{2}$) of all other pixels location at $(\Delta x, \Delta y)$ from the center, can be determined as follows:
%The GPS locations are extrapolated for every pixel of the weight map as in \eqref{lat} and \eqref{lon}. 

%\begin{equation}
\begin{align}
Lat_{2} &= Lat_{1} - \sin^{-1}( \frac{r \cdot \Delta y}{E_{r}})
\label{lat}\\
Lon_{2} &= Lon_{1} + \sin^{-1}( \frac{r \cdot \Delta x}{E_{r} \cdot \cos(Lat_{1} \cdot \frac{\pi}{180})}) \label{lon}
\end{align}
%\end{equation}
with $Lat_{1}$, $Lat_{2}$, $Lon_{1}$, $Lon_{2}$ representing latitudes and longitudes of the start and end points, $E_{r}$ the mean radius of the earth, $r$ the pixel ratio in meters per pixel depending on the ground altitude, and on the requested image scale and resolution, $\Delta x$ and  $\Delta y$ are the difference in pixels on the map between the two points.

\subsection{Global path planning} 
\begin{figure}[thpb]
	\centering
	\includegraphics[width=0.9\linewidth,height=\textheight,keepaspectratio]{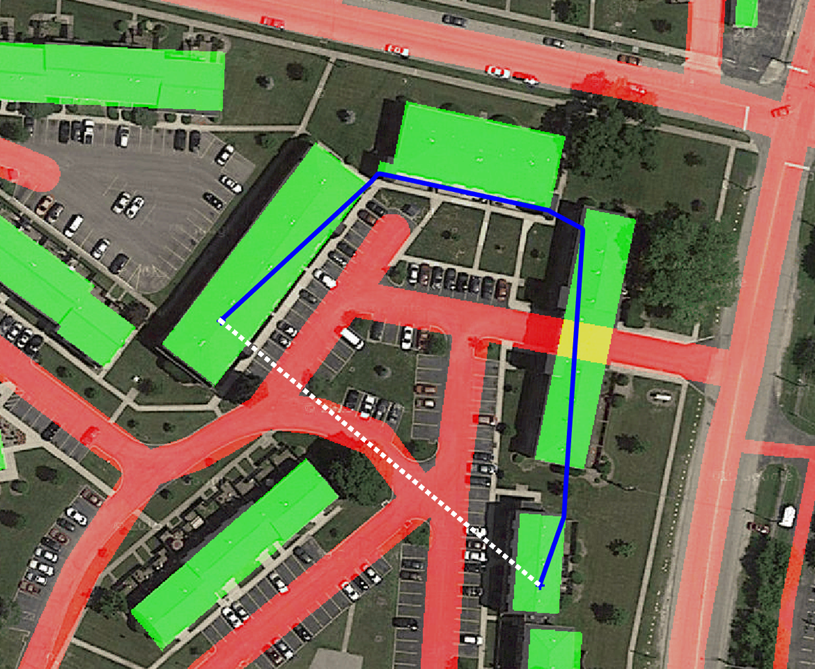}
	\caption{Different path planning solutions between two GPS coordinates. White dotted path: classic straight line path used in typical systems using waypoints. Blue path: our method, which determines the shortest path by minimizing the cost function such that the resulting path avoids flying over red areas that are dangerous and prefer green areas that are safer.\newline}
	\label{path}
	\includegraphics[width=0.9\linewidth,height=\textheight,keepaspectratio]{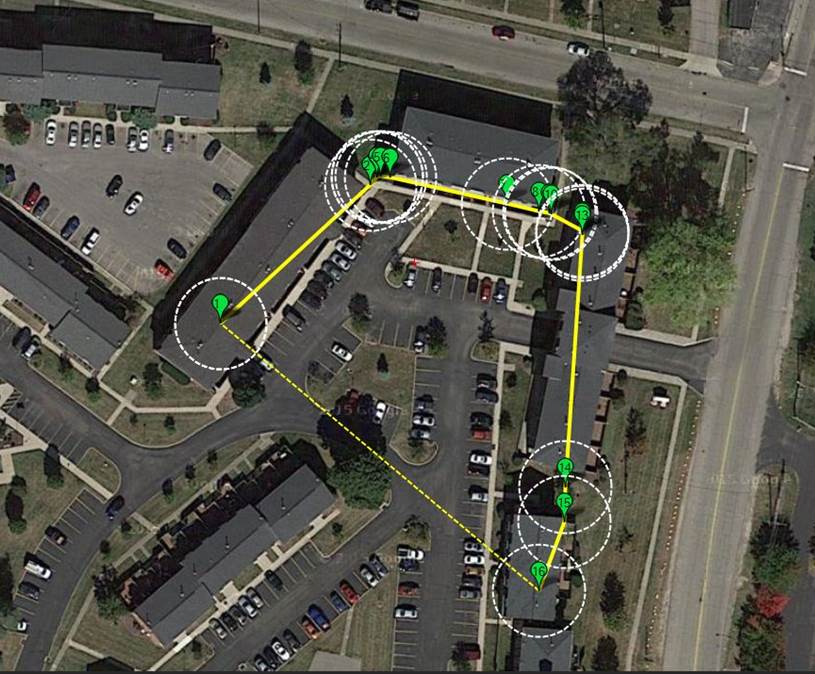}
	\caption{Path computed by algorithm, shown in yellow, converted to waypoints and visualized in Mission Planner. Green ticks are waypoints locations, white dotted circles are areas where the UAV will consider having reached the waypoint, and yellow dotted line represents the simple straight path.}
	\label{pathMP}
\end{figure}

The vast majority of UAS (Unmanned Aerial Systems) can be used with a ground control station (GCS), for example APM:Copter (previously known as ArduCopter) has its own mission planner with all the necessary tools. The conventional ways of controlling UAVs are either with a manual radio controller or by using a GCS that defines successive GPS waypoints (specifying the GPS location, altitude, and velocity) to which the UAV will fly autonomously. Despite their efficiency and convenience, there is a crucial flaw with waypoints; they are defined by the user and do not take into account the surroundings of the UAV. 
This is precisely what we want to tackle with our global path computation. By using the three types of data shown in Fig. \ref{layers}, we define the optimal path (example in Fig. \ref{path}) between two points and thus add a safety parameter to mission planning.

We find the safest route between two GPS coordinates by converting them into image pixels and computing a weighted shortest path algorithm using A* algorithm. The segments' lengths between two adjacent pixels are the Euclidian distance multiplied by the weight defined by the map class. Pixels that are in red have a weight of $100$, green is at $5$, and the rest is at $20$. Those values have been determined empirically. This process ensures that the red areas are avoided but also makes sure the UAV wouldn't take a long detour to reach its destination, thus keeping the loss of flight time to a minimum.

As the map is geo-registered, the outputted path can easily be converted into GPS coordinates using \eqref{lat} and \eqref{lon}, and put in KML an TXT files to be readable by mapping software and ground control stations (Fig. \ref{pathMP}).

This global weight map considers the static environment that the UAV will encounter such as roads and buildings. In order to ensure a higher level of safety in all stages of the flight, we also adapt the path locally during the flight in regards to moving objects as explained in the next section.

\subsection{Local path planning}
For increased safety, the path needs to be adapted dynamically during the flight to avoid moving objects detected in the field of view of the UAV's embedded camera.
In order to ensure a sufficient distance margin between each object and the UAV, the weight map used for shortest path is modified according to the objects' location, trajectory and velocity.
We compute the new weights of the map by applying at chosen locations a multivariate normal probability density function.

The variance $\Sigma x$ and $\Sigma y$ for each distribution are dependent on the object's characteristics. The $x$ term is proportional to the width of the object in pixels, and $y$ term \eqref{sig} is proportional to the object's velocity.
\begin{equation}
\Sigma y = V_{Obj} \cdot{S}
\label{sig}
\end{equation} 
where $V_{Obj}$ is the current velocity of the object and $S$ the safety margin to avoid collision in seconds.\newline

The resulting distribution is normalized, rotated to align with the object's trajectory, and centered on the chosen location (\ref{weightEqu}). The weight map is then multiplied with the distribution instead of being swapped in order to keep the global environment based information.

The locations where the distribution is applied to are defined given two criteria. One is whether the object collides with the UAV's path, and the other is how this collision happens. %based on if and how the object will collide with the UAV's path. 
The object and UAV will take respectively $t_{Obj}$ and $t_{UAV}$ seconds to the collision point, if $|t_{Obj} - t_{UAV}| < \Delta$ ($\Delta$ is set to 5 s in our experiments), 
%and if the object and UAV will cross with a delta in time $\delta t$ inferior to a given threshold $\Delta$ (5 seconds is used in our experiments). If a collision is detected and $\delta t < \Delta$, 
the distribution is applied on the collision point and also on a projected location to avoid re-planing a path that will create a similar situation. The projected location (\ref{T}, \ref{D}) is estimated as follows: the time for the UAV to travel to the current object's location is computed, the projected location is where the object will be at that time given constant velocity and trajectory for the object.
For objects that will not collide or that do meet the requirement of $\delta t > \Delta$, the distribution is applied at the next and projected locations.

\begin{equation}
\Omega = \Omega \cdot \left[ R \ \ T \right]  \circ \Phi\\
\label{weightEqu}
\end{equation} 
where $\Omega$ is the weight map, $\left[ R \ \ T \right]$ the affine transformation applied to the multivariate normal probability density function $ \Phi$ to allocate costs to $\Omega$. $ \Phi$ is centered at the wanted location $L_{x,y}$ and oriented given the rotation matrix $R$ dependent on the object's trajectory and

\begin{equation}
T = \begin{bmatrix}
L_{x}^{Obj} + D \cdot \sin (\alpha)\\
L_{y}^{Obj} + D \cdot \cos (\alpha)\\
1
\end{bmatrix}
\label{T}
\end{equation} 
is the translation component of the affine transformation, $L_{x}^{Obj}$ and $L_{x}^{Obj}$ are the image coordinates of the object, and $D \cdot \sin (\alpha)$ and $D \cdot \cos (\alpha)$ are respectively the distances in $X$ and $Y$ to the desired location.

\begin{equation}
D =
\begin{cases}
V_{Obj} \cdot \frac{d_{UAV-Obj}}{V_{UAV}} \text{ for projected location}\\
V_{Obj} \cdot \frac{1}{fps} \text{ \hspace{0.305in} for next location}
\end{cases}
\label{D}
\end{equation} 
with $D$ the distance used in \ref{T}, $V_{Obj}$ and ${V_{UAV}}$ the current velocities of the object and UAV, $d_{UAV-Obj}$ the distance between object and UAV, and $fps$ the frame-rate or computation time.\\

\begin{figure}[thpb]
	\centering
	\includegraphics[width=0.9\linewidth,height=\textheight,keepaspectratio]{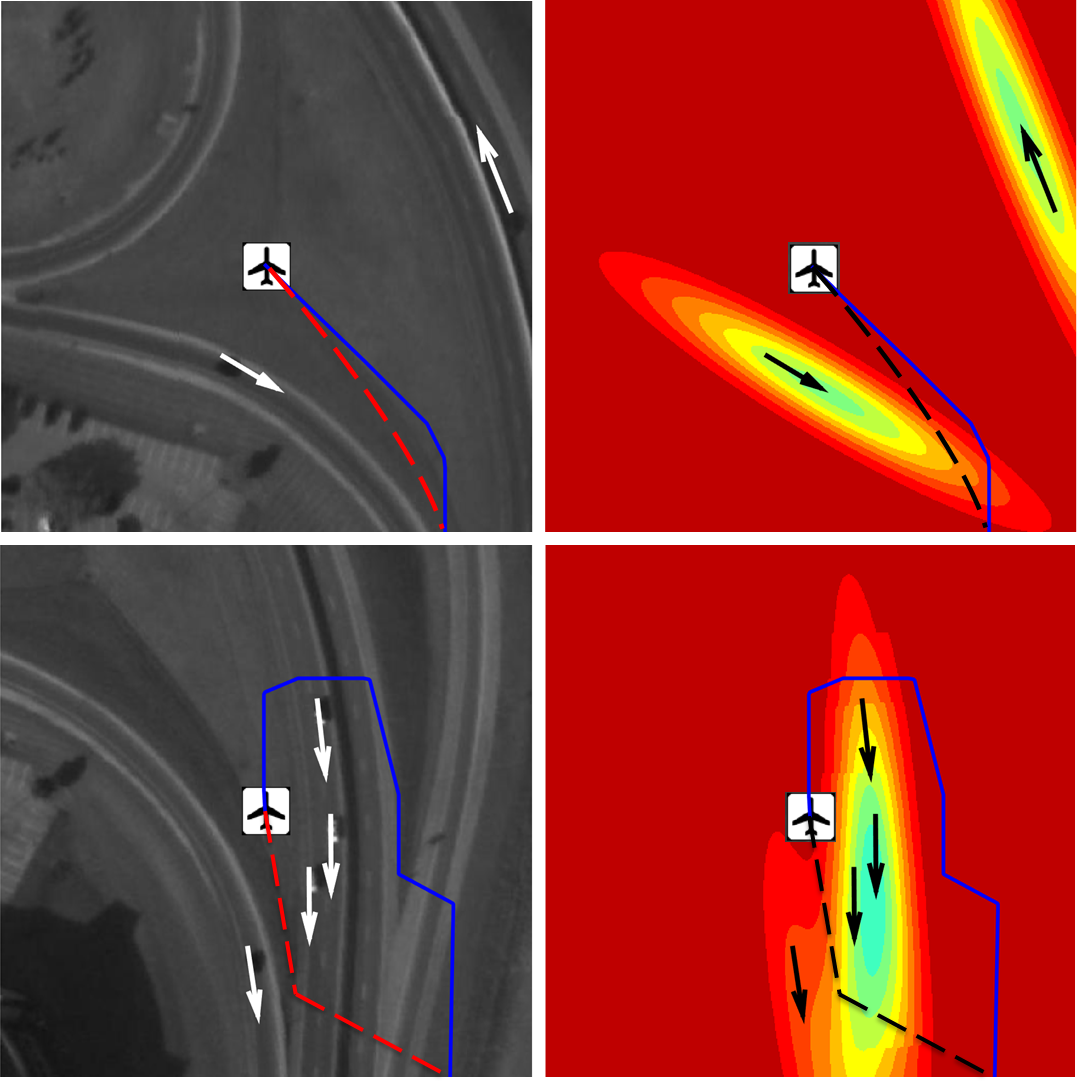}
	\caption{
		Left column shows the images and right column shows the corresponding weight maps. Objects trajectories are shown in white (in images) and black (in weight maps). The global paths	shown in red and black dashed lines, are adapted with weight adjustments to avoid flying over objects. 
		Please note that, in the bottom row, the path crosses the road perpendicularly on the right part of the images (more visible on the bottom left image). This is due to the fact that we want to minimize crossing high-cost road pixels.
		%Map weight change for moving objects in the FOV. The global path, shown in red and black dashed lines, is adapted with weight adjustments to avoid objects. Left and right images, respectively dataset image and weight map, are overlaid with new path in blue. Objects' location and trajectory are shown with white and black arrows. Please note in the bottom row how the path crosses the road perpendicularly on the right part of the images, it proves that the initial weights are still present.
		}
	\label{avoid}
\end{figure}

This method will ensure that the resulting path will leave sufficient ground distance between objects and the UAV, and if multiple objects are close together, it will create a barrier and encourage the UAV to find a safer path, thus preventing it to fly above any moving objects (Fig. \ref{avoid}).

%%%%%%%%%%%%%%%%%%%%%%%%%%%%%%%%%%%%%%%%%%%%%%%%%%%%%%%%%%%%%%%%%%%%%%%%%%%%%%%%
\section{EXPERIMENTS}
\subsection{Methodology}
In order to simulate a real world scenario as accurately as possible, our method uses dataset images, and typical UAVs' specifications and camera parameters. We made sure to comply with the latest regulations and advice regarding UAV operation, and used the following flight and hardware parameters:
\begin{itemize}
	\item Altitude above ground level : $50 m$. 
	\item Velocity : $< 15 m/s$.
	\item Camera Horizontal Field Of View (HFOV) : $97.40^{o}$ \footnote{HFOV for a configuration using a PointGrey Blackfly 1,3MP 1/3" camera of 1288x964 resolution and a Kowa LM3PB lens.}.
	\item Horizontal ground sampling resolution : $8.84 cm/pixel$.\\
\end{itemize}

\begin{figure}[thpb]
	\centering
	\includegraphics[width=0.95\linewidth,height=\textheight,keepaspectratio]{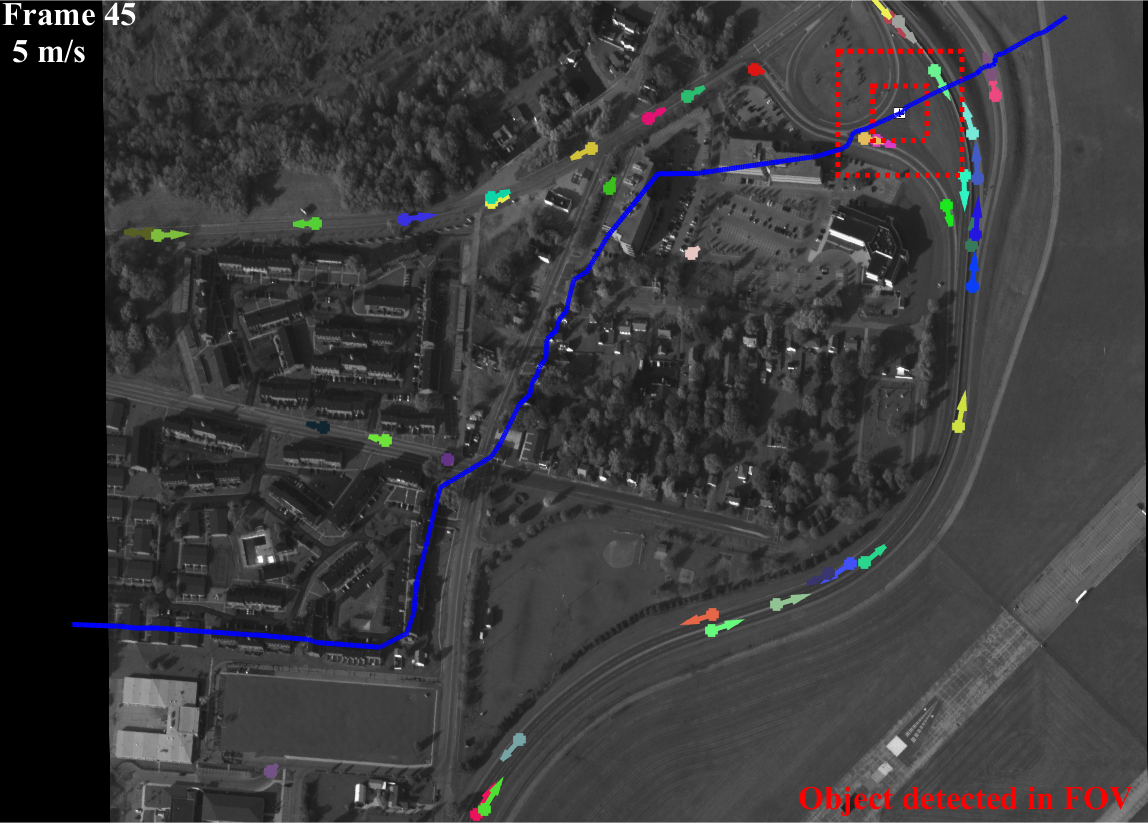}
	\caption{Small red dotted square on the top right represents the danger area that the UAV would reach in 5 seconds at the current	velocity, and the larger square shows the FOV. Objects’ locations and their motion vectors given by ground truth are shown by colored arrows. At bottom right a notification is displayed in red if objects are present in the FOV.
		%Interface visible during the simulation process. Top left: number of the displayed frame and current velocity of UAV. Blue line: global path. White plane icon: current location of the UAV. Red dotted lines: the smaller represents the danger area that the UAV would reach in 5 seconds at current velocity, and the bigger represents the FOV. Bottom right: Notification if objects are in the FOV. Color square and arrows: objects' locations and motion vectors given by ground truth.
		}
	\label{snap}
\end{figure}

The principles used to build the simulation scheme are the following: 
\begin{itemize}
	\item UAV videos are registered in the geo-referenced space, we can thus work in pixels coordinates, and convert back to GPS anytime. 
	\item The datasets' ground truth gives the moving objects' location for every frame (motion vectors in Fig. \ref{snap}). 
	\item The UAV will follow the global path (blue in Fig. \ref{snap}). 
	\item For every frame the UAV's displacement in the image is dependent on it's velocity and direction \eqref{dp}.
	\item The considered objects are only the ones visible in the field of view of the embedded camera (exterior red dotted line around the UAV in Fig. \ref{snap}).
	\item For convenience we call \textquoteleft collision \textquoteright the situation where the UAV will fly over an object.
	\item A collision is detected if the direction of an object's motion vector intersects the path in front of the UAV.
	\item A danger area is computed, and is visible as the smallest red dotted rectangle in Fig. \ref{snap}, for every frame depending on UAV's velocity so that the UAV will reach the boundary in 5 seconds at current and constant velocity.
\end{itemize}

\begin{equation}
\Delta p=\frac{V_{UAV}}{F_{d}\cdot r_{m}}
\label{dp}
\end{equation}
where $\Delta p$ is the number of pixels to advance along the path, $V_{UAV}$ is the velocity of the UAV, $F_{d}$ is the framerate of the dataset, and $r_{m}$ represents the ground sampling distance of the geo-registered map.\newline

\subsection{Datasets}
We use two datasets to run our safe navigation pipeline, Wright-Patterson Air Force Base (\href{https://www.sdms.afrl.af.mil/index.php?collection=wpafb2009}{WPAFB}) \cite{WPAFB} and \href{http://www.pv-labs.com/}{PVLabs}. They are wide-area motion imagery (WAMI), and provide ground truth for moving objects on ortho-rectified images captured by UAVs. Both of those datasets have been captured at high altitude with embedded sensors and a matrix of multiple cameras. We use the provided regions of interest outputted by a geo-registration step, described in \cite{WPAFB}. 
For each dataset we run the different steps of the pipeline. We first create the weight map using the process described in section \ref*{sec3A}. Videos are then precisely geo-registered onto the map via homography transformation. The global path (Fig. \ref{paths1}) is generated before the simulated flight and adapted dynamically on the way.

%%%%%%%%%%%%%%%%%%%%%%%%%%%%%%%%%%%%%%%%%%%%%%%%%%%%%%%%%%%%%%%%%%%%%%%%%%%%%%%%
\section{RESULTS}

\begin{figure}[thpb]
	\centering
	\includegraphics[width=0.85\linewidth,height=\textheight,keepaspectratio]{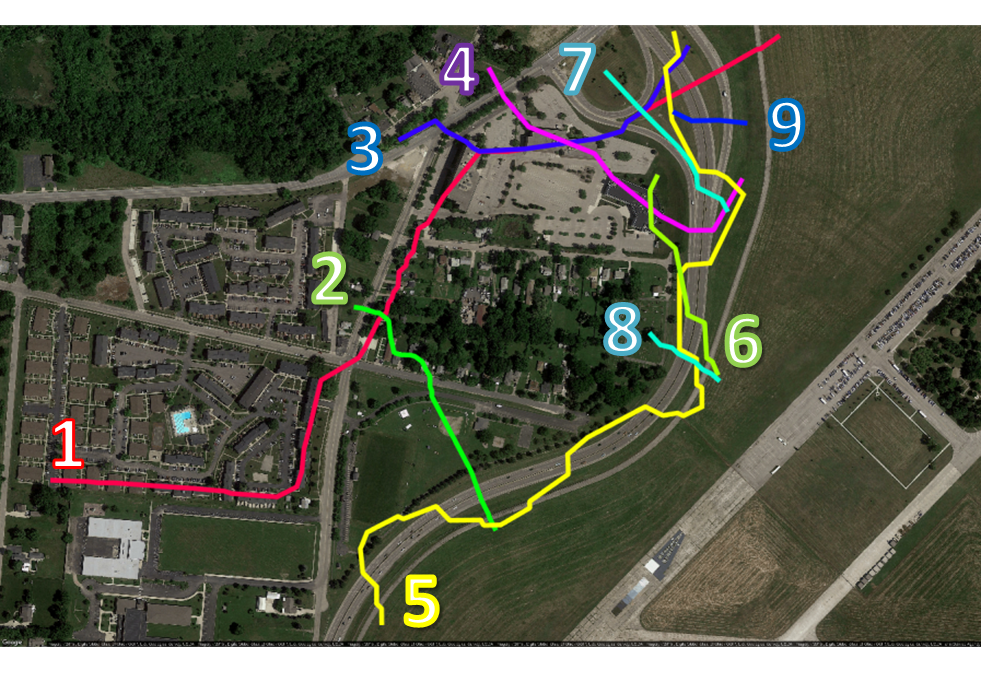}
	\includegraphics[width=0.65\linewidth,height=\textheight,keepaspectratio]{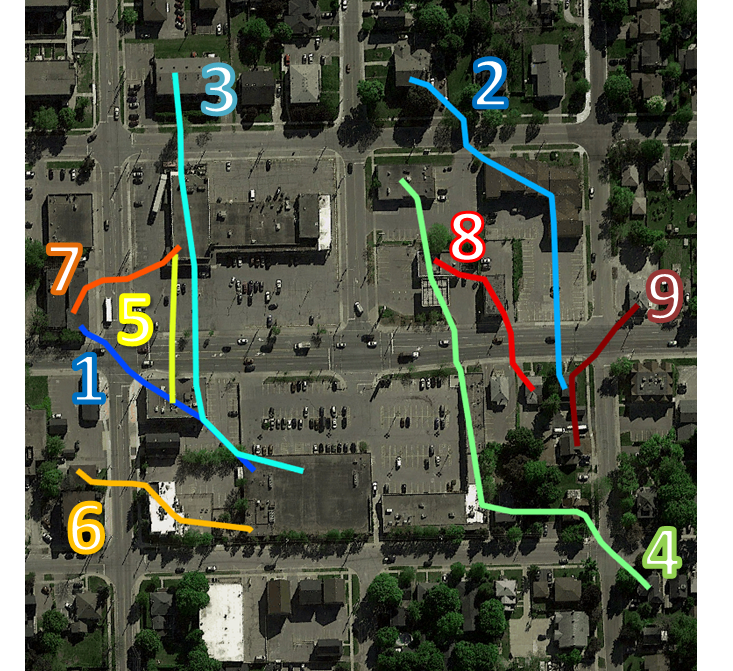}
	\caption{Visualization of the nine paths that have been tested (each at 5, 8, and 11 m/s) for both datasets. Top: WPAFB. Bottom: PVLabs. Images acquired using the Google Maps API.}
	\label{paths1}
\end{figure}

For both WPAFB and PVLabs datasets, we defined 9 different pairs of start and end GPS coordinates (Fig. \ref{paths1}) based on the environment and busyness of the roads to create challenging situations that will require global path adaptation. And each path is executed at three different UAV velocities: 5, 8, and 11 m/s.
The total traveled distance by using the global path compared to the classic straight line path, for each dataset executed for all paths at three above velocities, is $20\%$ higher or 5:13 min longer for WAPAFB and $6\%$ or 32 s for PVLabs, making our safety increased path an affordable measure in term of autonomy.

To quantify the performance of the proposed method we introduce a metric assimilated to safety. We consider the UAV to object proximity, the closer the UAV is to an object the more danger it represents for it, we therefore compute a total cost for each dataset as in \eqref{cg}.

\begin{equation}
C_{g} =\sum_{i=1}^{\#Objects} \alpha \cdot e^{-D_{ou}^{i}}
\label{cg}
\end{equation} 
with $C_{g}$ the global cost for the considered dataset, $D_{ou}$ the ground distance between the UAV and each object detected in the FOV during the experiment, and $\alpha$ a constant.\newline 

Note that, for the same start and end locations, when different paths are compared, the UAV will not encounter the same situations. This is why, for clarity, we include, with the results in Table \ref*{tabRes}, the number of objects seen by the UAV's camera throughout the simulation for each dataset.

\begin{table}[htpb]
	\caption{Safety estimation results for WPAFB and PVLabs}
	\label{tabRes}
	\begin{center}
		\begin{tabular}{|c|c|c|c|}
			\hline
			 & Straight path & Static path & Dynamic path \\
			\hline
			WPAFB \# of obj. & \textbf{2,759} & 4588 & 7,597 \\
			\hline
			Global WPAFB cost & 243.9 & 62.9 & \textbf{5.6} \\
			\hline
			PVLabs \# of obj.  & \textbf{3,600} & 4,022 & 5,959 \\
			\hline
			Global PVLabs cost & 188.1 & 326.3 & \textbf{98} \\
			\hline
		\end{tabular}
	\end{center}
\end{table} 

\begin{figure}[thpb]
	\centering
	\includegraphics[width=0.49\linewidth,height=\textheight,keepaspectratio]{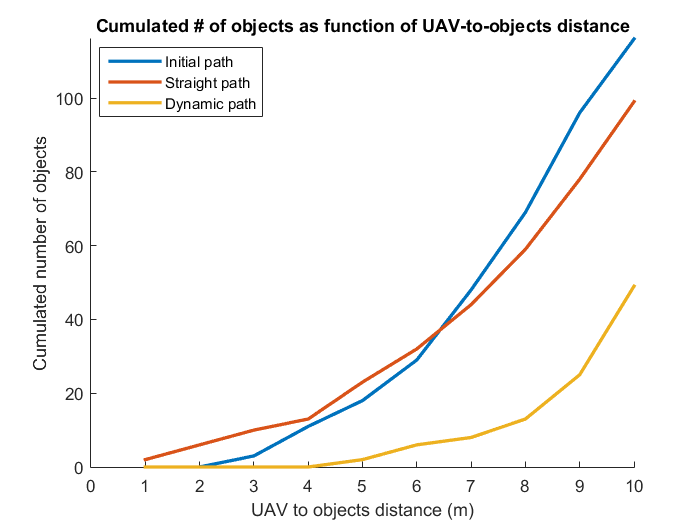}
	\includegraphics[width=0.49\linewidth,height=\textheight,keepaspectratio]{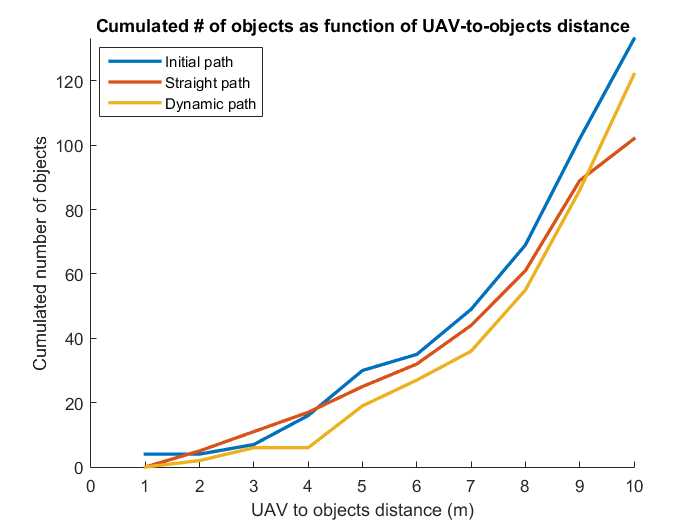}
	\caption{Comparison of the number of detected objects in the FOV as function of UAV-to-objects ground distance between 0 and 10m for all nine paths executed at 5, 8, and 11 m/s. The perfect solution would be 0 objects for all distances. Left: WPAFB. Right: PVLabs.}
	\label{curves1}
\end{figure}

We can clearly see in Table \ref*{tabRes} that our proposed method encounters more objects in the FOV, but it has the means to keep the UAV afar from them. Objects which are over 20m away are not in danger, but having a car or pedestrian closer than 5m to the UAV represents a very concerning situation in terms of safety for people. This is why we have chosen to compute the global cost with a negative exponential weight function, that way the shorter the distance, the more cost is applied to the global metric. The proposed method encounters over twice the amount of moving objects but safely keeps away from them (Fig. \ref{curves1}), making the resulting safety parameter much better than global path and, most of all, better than classic straight line path.

%%%%%%%%%%%%%%%%%%%%%%%%%%%%%%%%%%%%%%%%%%%%%%%%%%%%%%%%%%%%%%%%%%%%%%%%%%%%%%%%
\section{CONCLUSION}
In this paper we introduced an environment and safety based path planning for low altitude UAV operating in urban areas. We compute a global path, for any mission given a pair of start and end GPS locations, by using a weighted shortest path. The weight map is defined using ground classification data summarized in three classes: highest cost is for roads and paths because of the high probability of presence of people for which the UAV represents a safety threat, safest are buildings and water, and neutral areas are the rest. Additionally, we included a dynamic path planning that will modify locally the flight plan while in flight to avoid being close to moving objects such as vehicles and pedestrians. Our proposed method has been tested in simulation using geo-registered data and images from two WAMI datasets, WPAFB and PVLabs, and it showed significant improvement compared to the current and manual mission planning solution in terms of a safety metric quantifying threat in function of UAV-to-object distance.

Our safety planning and navigation scheme can be implemented on-board a UAV and will consist in the following steps: 1- before takeoff, acquire necessary GIS data for the mission area, and generate mission waypoints using global weighted path planning, 2- during the flight, geo-register the embedded camera's images using a sensor model and gimbal readings, detect moving objects (as in \cite{tca}) or any other type of objects to avoid, and generate new local path and waypoints to stay clear of the detected objects.

\addtolength{\textheight}{-12cm}   % This command serves to balance the column lengths

%%%%%%%%%%%%%%%%%%%%%%%%%%%%%%%%%%%%%%%%%%%%%%%%%%%%%%%%%%%%%%%%%%%%%%%%%%%%%%%%
\section*{ACKNOWLEDGMENT}
The research was supported by a DGA-MRIS scholarship.

%%%%%%%%%%%%%%%%%%%%%%%%%%%%%%%%%%%%%%%%%%%%%%%%%%%%%%%%%%%%%%%%%%%%%%%%%%%%%%%%

\end{document}